\documentclass[sigconf, nonacm]{acmart}
\usepackage{algorithm}
\usepackage{algpseudocode}
\usepackage{graphics}
\usepackage{soul}
\usepackage{float}

\AtBeginDocument{%
  }

\setcopyright{acmlicensed}
\copyrightyear{2026}
\acmYear{2026}
\acmDOI{XXXXXXX.XXXXXXX}

\acmConference[XXX '26]{XXX}{XXX, 2026}{XXX}
\acmISBN{978-1-4503-XXXX-X/18/06}

\graphicspath{{figures/}}

\begin{document}

\title{LiDDA: Data Driven Attribution at LinkedIn}


\author{John Bencina}
\authornote{These authors contributed equally}
\affiliation{%
\institution{LinkedIn Corporation}
\city{Sunnyvale}
\country{USA}}
\email{jbencina@linkedin.com}

\author{Erkut Aykutlug}
\authornotemark[1]
\affiliation{%
\institution{LinkedIn Corporation}
\city{Sunnyvale}
\country{USA}}
\email{eaykutlug@linkedin.com}

\author{Yue Chen}
\affiliation{%
\institution{LinkedIn Corporation}
\city{Sunnyvale}
\country{USA}}
\email{florachen0014@gmail.com}

\author{Zerui Zhang}
\affiliation{%
\institution{LinkedIn Corporation}
\city{Sunnyvale}
\country{USA}}
\email{zerzhang@linkedin.com}

\author{Stephanie Sorenson}
\affiliation{%
\institution{LinkedIn Corporation}
\city{Sunnyvale}
\country{USA}}
\email{ssorenson@linkedin.com}

\author{Shao Tang}
\affiliation{%
\institution{LinkedIn Corporation}
\city{Sunnyvale}
\country{USA}}
\email{tangshao28@gmail.com}

\author{Changshuai Wei}
\authornote{Corresponding author}
\affiliation{%
\institution{LinkedIn Corporation}
\city{Seattle}
\country{USA}}
\email{chawei@linkedin.com}

\renewcommand{\shortauthors}{Bencina et al.}

\begin{abstract}
Data Driven Attribution, which assigns conversion credits to marketing interactions based on causal patterns learned from data, is the foundation of modern marketing intelligence and vital to any marketing business and advertising platform. In this paper, we introduce a unified transformer-based attribution approach that can handle member-level data, aggregate-level data, and integration of external macro factors. We detail the large scale implementation of the approach at LinkedIn, showcasing significant impact. We also share learnings and insights which are broadly applicable to the marketing and ad tech fields. 
\end{abstract}

\ccsdesc[500]{Computing methodologies~Artificial intelligence}
\ccsdesc[500]{Applied computing~Marketing}
\ccsdesc[500]{Information systems~Online advertising}
\ccsdesc[300]{Mathematics of computing~Probability and statistics}
\keywords{Multi-Touch Attribution, Media-Mix Modeling, Attention Mechanism, Computational Marketing, Transformer}


\maketitle
\sloppy

\section{Introduction}
Marketers rely on attribution measurement to understand the impact of marketing campaigns on customers' journeys and decisions. Accurate attribution enables marketers to better optimize marketing spend and strategy to maximize their return on investment (ROI). In this paper, we propose a framework for large scale end-to-end attribution system and detail the implementation at LinkedIn. LinkedIn Data Driven Attribution (LiDDA) leverages a combination of bottom-up and top-down modeling approaches in order to build a more accurate attribution platform to measure campaign performance and ROI. The system was originally developed for our internal Go-To-Market (GTM) Marketing and has since expanded to support external applications for LinkedIn customers.

\subsection{LinkedIn Use Case}

GTM Marketing is responsible for managing digital marketing campaigns for various LinkedIn products (e.g., Recruiter Lite, Sales Navigator), targeting both acquisition and retention objectives, where marketing audiences include both consumers and buyers. Consumer marketing focuses on the individual as a member such as member account signups. Buyer marketing focuses on the company-level conversions such as the use of LinkedIn Ads platform ads or Talent Solutions job posts.

Both segments are targeted using a combination of owned and external channels. Owned channels include Email and the LinkedIn Ads Platform (Feed Ads) where we own the platform used and have greater visibility to individual engagements. External channels refer to other online platforms such as Paid Search, Social Media, and Streaming. These channels do not provide individual level data, and we are often limited to aggregate information such as the total number of impressions in a time period.

One major challenge is lack of a unified view of the member journey across multiple channels. Existing media-mix-modeling (MMM) approaches provide channel-level visibility; however they lack detailed campaign breakouts and journey information. Last touch attribution (LTA) was the default standard; yet it biased results to bottom of funnel (BOFU) campaigns and omitted channels without individual identifiers.

Through our work on LiDDA, we are able to build an approach which combines signals using both top-down and bottom-up information. This enables our marketing team to utilize attribution across different levels of granularity while still maintaining consistency. In the next section we further discuss the differences in top-down and bottom-up attribution approaches.

\subsection{Overview of Related Work}

Data-driven attribution (DDA) aims to measure the contribution of online campaigns at the user level using a bottom-up approach. On the other hand, Media-Mix modeling (MMM) considers both online and offline channels along with seasonality using a top-down (aggregate time-series) approach.

\subsubsection{Media-Mix Modeling}
Media-mix models (MMM) use aggregated time-series data to model conversions as a function of online and offline channels as well as macro-economic factors and seasonality. MMM have been widely used since the 1960s \cite{borden64, mcCarthy78}, and provide a holistic view across all channels and external factors. On the other hand, they come with different challenges \cite{chenPerry2017}. One of the shortcomings is that they do not have visibility to analyze data at the user level.  

\subsubsection{Rules Based Attribution}
Traditional attribution models commonly use rule-based attribution (RBA) approaches such as first-touch, last-touch, or time decay\cite{gaur2020attribution}. These approaches assign conversion credit based on simple, predetermined rules at user level. RBA methods are easy to understand and implement; however, they are often based on limited assumptions and introduce biases.

\subsubsection{Data-Driven Attribution}
\label{subsection:relatedDDA}
Data-driven attribution (DDA) addresses the limitations of RBA by leveraging machine learning and statistical techniques to allocate conversion credit learned from data. Different modeling approaches such as Logistic regression \cite{shao2011data}, Bayesian \cite{liKanan14, xuetal14} and Survival Theory \cite{zhang2014multi} have been used. Deep learning approaches such as CAMTA \cite{kumar2020camta}, DNAMTA\cite{li2018deepneuralnetattention}, and CausalMTA \cite{yao2022causalmta} have demonstrated improved performance with these classes of models.

\subsection{Our contributions}
As far as we know, LiDDA is the first industry-deployed large-scale DDA system that leverages transformer-type attention mechanism and can handle member-level, aggregate-level and  integration
of external macro factors. In particular, the system is highlighted by:
(i) \textit{Transformer Architecture}: our use of transformer type of attention offers
both single-pass inference and an explainability mechanism; (ii) \textit{Temporal-aware attention}: we discretize time gaps between campaign touchpoints and learn positional embeddings—plus a special day‑of‑week embedding—to capture both sequence order and seasonal effects; (iii) \textit{Privacy‑driven data imputation}: to correct for missing paid‑media signals (e.g. due to GDPR/CCPA restrictions), we probabilistically inject aggregated impressions back into user paths; (iv) \textit{Unified DDA\&MMM outputs}: we reconcile discrepancies between independent DDA and macro‑level MMM by training-time calibration; (v) \textit{End‑to‑end solution}: we present our modeling innovations, implementation details, comprehensive evaluation and insights from launching to both GTM marketing and Ads platform in the paper.

\section{LiDDA System}
\subsection{System Overview}
\begin{figure}
    \centering
    \includegraphics[width=1\linewidth]{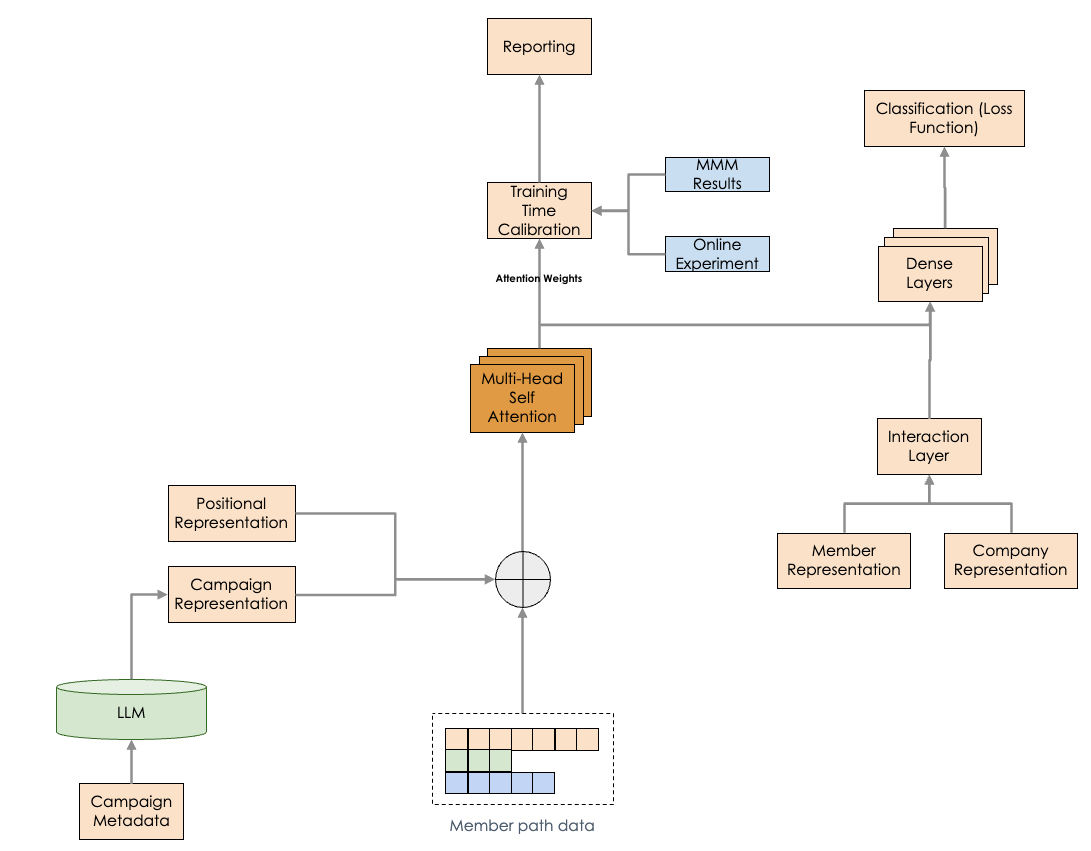}
    \caption{High level diagram showing the end-to-end components of our attribution modeling framework}
    \label{fig:lidda_arch}
    \Description{}
\end{figure}

LiDDA is our modeled attribution platform (Figure ~\ref{fig:lidda_arch}) which utilizes an attention-based modeling approach. Attention models are well suited for this task since they retain sequential information about the buyer journey and also enable us to incorporate additional features about the member, their company, and campaign information. 

Overall, the model is structured as a binary classification task to predict whether a buyer journey will result in a conversion 
$P(Y | E_{M}, E_{C}, S),$
where \(Y\) denotes the binary conversion outcome, \(E_{M}\) denotes the member embedding with length $D_M$, \(E_{C}\) denotes the (member’s) company embedding with length $D_C$, and $S$ denotes the sequence of marketing touchpoints on the member with length $N$.
\(E_{M}\) is a member embedding derived from LinkedIn data such as skills, titles, etc. that generally capture latent features of that member. Likewise, \(E_{C}\) is a company embedding derived from LinkedIn data such as industry, roles, etc.

From the trained model, we output the aggregated attention weight matrix from the touchpoint sequence \(S\). The attention weights are normalized such that they add up to 1.0, i.e., $\hat{\alpha}_i = \frac{\alpha_i}{\sum_{j=1}^{n} \alpha_j},$
and the normalized weights are interpreted as percent contribution of that touchpoint to the final conversion event.  We further discuss attention in Section \ref{subsection:attention}.

The attention weights represent the relative credits within the path; however they still need to be rescaled for overall impact. We use a separate media-mix model (MMM) to adjust the DDA credits at the channel level to account for overall incrementality and channels where individual exposure data does not exist or is not available. This is discussed further in Section \ref{section:outputScaling}.

\subsection{Path Creation}

The touchpoint sequence $S$ is modeled as a time-ordered series of interactions, where each touchpoint,  $t \in \mathcal{T}$, is one of the possible (channel, action) pairs, such as (EMAIL, OPEN) or (AD, CLICK). Sequences are truncated to the most recent $N$ interactions, and each touchpoint is encoded as a learned embedding capturing its latent characteristics. These embeddings are combined across time to form an $N \times d$ representation of the customer journey.

To enrich the representation, each timestep also includes a marketing campaign embedding $E_{MCID}$ $\mathbb{R}^{MCID \times {\text{dim}}}$, generated from campaign metadata and content using an upstream LLM fine-tuned on LinkedIn data. This approach allows flexibility in combining structured and unstructured data consistently across domains.

\subsection{Positional Representations}
\label{subsection:position}
\subsubsection{Positional Encodings}
Attention models do not inherently model the order of touches in the paths so additional encodings are needed to capture the relative positions of each touch. Sinusoidal positional encodings are commonly used for temporal ordering as in the original attention design \cite{vaswani2017attention}. These are implemented as: $PE_{(pos,2i)} = sin(pos / 10000^{2i/d_{model}})$, $
    PE_{(pos,2i+1)} = cos(pos / 10000^{2i/d_{model}})$


Because our touchpoint embedding \(E_{T}\) is at the granularity of (channel, action), the number of possible combinations is far lower than typical NLP vocabulary sizes. Therefore we use a substantially smaller embedding dimension for our data. We employ Time Absolute Position Encodings (tAPE) where the positional encoding above is modified to adjust for the lower dimensional representation\cite{Foumani_2023}. The tAPE encodings are defined as:
$\omega_k=10000^{-2k/d_{model}}$, $\omega_k^{new} = \frac{\omega_k\times d_{model}}{L}$.


\subsubsection{Positional Embeddings}
Positional encodings are not sufficient for our domain where marketing engagements are generated with irregular patterns and not across fixed intervals. For example, the time between an impression and the following click may be a few seconds while the time between two impressions could be days. We capture these irregular differences by representing the time difference between interactions as a discrete interval. For LinkedIn use cases, we use days as the interval unit since the look-back-window is limited to a few months based on business requirements.

Each touchpoint has a time interval relative to the final conversion state in the path, encoded by days as $D$ with a corresponding learned embedding $E_D$. Due to the strong seasonality of our marketing data, we also learn embeddings \(E_{DOW}\) corresponding to DOW encoding of each touchpoint.

We combine all the embeddings together for each touchpoint as: $concat(E_{MCID}, E_{T}) + E_{D} + {E_{DOW}} + tAPE$.


\subsection{Attention Mechanism}
\label{subsection:attention}
We explored different methodologies for performing data driven attribution in our model as discussed in Section \ref{subsection:relatedDDA}. In addition to their efficiency, models incorporating attention layers such as DNAMTA and CAMTA \cite{li2018deepneuralnetattention, kumar2020camta}, outperformed other model designs in this domain. Attention was originally proposed for NLP tasks \cite{vaswani2017attention}, but later adopted in many other domains. Formally, attention is defined as:

\begin{equation}
  \mathrm{Attention}(Q, K, V) = \mathrm{softmax}(\frac{QK^T}{\sqrt{d_k}})V
\end{equation}

We utilize the multi-head variation of attention where each head $H$ has the ability to generate its own attention weights:
\begin{equation}
\mathrm{MultiHead}(Q, K, V) = \mathrm{Concat}(\mathrm{h}_1, \mathrm{h}_2, \dots, \mathrm{h}_n)
\end{equation}

The attention mechanism enables the model to adjust the weight given to different tokens in a sequence to optimally solve its objective. In LiDDA, we utilize a self-attention setup where \(Q\), \(K\), and \(V\) are equivalently the marketing engagements performed by the member. We apply attention over the sequence $S$ and use the attention scores in the training classification task. The attention operation produces a representation of the sequence $E_S$ where $E_S \in \mathbb{R}^{N \times H \times {\text{dim}}}$. We aggregate the heads by averaging the scores across the head dimension to obtain $E'_S \in \mathbb{R}^{N \times {\text{dim}}}$.

Different strategies exist for performing classification from attention scores. In BERT \cite{devlin2019bertpretrainingdeepbidirectional}, the authors created a special [CLS] token at the start and use only the learned representation of the first token. Other valid approaches include max-pooling or mean-pooling over the entire sequence. During inference, we compute the final attribution output as the summation of the attention weights across all heads and then normalize each row to 1.0 such that each row represents a path and each value in that row represents the \(j\)-th touchpoint in that sequence. We set masked values to 0 if a path is shorter than our maximum length.

In LiDDA, we keep all attention scores by flattening the attention output. The motivation behind this decision is to retain as much positional information as possible when computing the attention weights. Additionally, we limit our attention module to a single layer in order to reduce behavior where attention interpretability may be affected by abstract representations across timesteps as discussed in the next section.

\subsubsection{Attention for Attribution}
\label{subsection:relatedAttention}
Prior works \cite{li2018deepneuralnetattention, kumar2020camta} utilize attention-based crediting by applying a self-attention layer over a recurrent network. Our initial exploration of this approach found a heavy bias towards end-of-path touches. Using a bi-directional LSTM may help alleviate this effect; however we modified the approach to directly apply self attention over the input sequence. One issue with the LSTM encoder is that the hidden state contains information about previous timesteps, making attention interpretation less straight forward \cite{serrano2019attention}. The topic of attention interpretability has been the subject of extensive research \cite{jain-wallace-2019-attention, wiegreffe2019attention}. The central concern is that the resulting weights may not faithfully align with the model final output. While our modification can still introduce interpretability issues in attention based models \cite{brunner2020identifiability}, we observe within our domain that attention weights do impact model output.

It is also possible to compute attribution based on the incremental propensity from the prior engagement \cite{shao2011data} or as an approximation of Shapley values \cite{yao2022causalmta}. These methods are not necessarily superior. Counterfactual simulations can be computationally expensive and require accurately determining the counterfactual which is non-trivial. We did evaluate the attributions under the attention vs incremental propensity and found negligible differences (Table \ref{tab:method-weights}). We also observed that incremental propensity approach often produces early jumps causing front-loaded attribution as changes in propensity saturate with length. 
\begin{table}
    \centering
    \begin{tabular}{ccc}
        \toprule
        Channel & Incremental & Attention \\
        \midrule
        Channel A & 14.4\% &  12.7\% \\
        Channel B & 52.3\% &  53.4\% \\
        Channel C & 29.5\% & 32.5\% \\
        Channel D & 3.8\% & 1.5\% \\
        \bottomrule
    \end{tabular}
    \caption{Comparison of incremental propensity and attention weight from LiDDA model.}
    \label{tab:method-weights}
\end{table}

\subsection{Imputation of Paid Media Touchpoints}
\label{section:paidMediaImputation}
LinkedIn's outbound marketing channels are grouped into owned and paid channels. For owned channels, we can collect touchpoints from all actions at the user level, e.g., sends, opens, clicks, impressions, while for external channels, impression data at the user level are not available, e.g., due to privacy regulations like GDPR or CCPA. 

\begin{algorithm}[h]
\caption{Allocate Member Impressions}
\label{algo:imputation}
\begin{algorithmic}
\State \textbf{Inputs:} 
\State \quad External total Imp. and Clicks $\{I^{\rm ext}_{c,d},C^{\rm ext}_{c,d}\}$, clicks set $J_{i,c,d}$
\State \quad Owned Impressions $\{I^{\rm own}_{\rm total}, I^{\rm own}_{\rm clickers}\}$
\State \quad No-click paths $\mathcal P_d$, path‐weights $\{u_k\}$
\medskip
\State \textbf{Outputs:} click‐imps $\{m_j\}$, path‐imps $\{I_k\}$
\medskip

\State {\bfseries Step 1: Member‐only budget}
\[
I_d =\sum_c I^{\rm ext}_{c,d}\;\frac{\sum_i|J_{i,c,d}|}{C^{\rm ext}_{c,d}}
\]
\State {\bfseries Step 2: Stochastic click allocation}
\[
\alpha^* = \frac{I^{\rm own}_{\rm clickers}}
               {I^{\rm own}_{\rm total}},
\quad
r_{c,d} = \alpha^*\,\frac{I^{\rm ext}_{c,d}}{C^{\rm ext}_{c,d}}
\]
\[
m_j \;\sim\;\mathrm{Geom}\bigl(p=\tfrac1{r_{c(j),d}}\bigr),
\quad
L_d = I_d - \sum_j m_j
\]
\State {\bfseries Step 3: Stochastic leftover allocation}
\[
p_k = \frac{u_k}{\sum_{\ell\in\mathcal P_d}u_\ell},
\quad
\lambda_k = L_d\,p_k,
\quad
\tilde I_k\sim\mathrm{Poisson}(\lambda_k)
\]
\[
I_k = \Bigl\lfloor \tilde I_k\,\frac{L_d}{\sum_{\ell}\tilde I_\ell}\Bigr\rfloor
\]
\end{algorithmic}
\end{algorithm}

When a user clicks on an ad on an external platform they land on a LinkedIn microsite. We can use the URLs from microsites as a proxy for paid clicks at the user level. Meanwhile, we can also obtain daily aggregate campaign-level data, e.g, number of impressions or clicks for a campaign on a given day. Together with member-level clicks and impressions from owned channel, we can impute paid media impressions probabilistically.

Let $i$ index members, $d$ days, $c$ channels, $j\in J_{i,c,d}$ external‐click events, and $k\in\mathcal P_d$ no‐click paths (i.e., path without clicks from external channel).  We perform the imputation in three steps (Algorithm~\ref{algo:imputation}): (i) Estimate member impression from external total impressions, yielding total "budget" that needs allocation;
(ii) Compute impression per click among the clickers $r_{c,d}$ and then draw impression from corresponding geometric distribution;
(iii) Assign weights $u_k$ for no-click path $k$ based on campaign and location information, then draw Poisson counts of impressions for no-click path to allocate remaining impressions. Further information, including the step-by-step procedure, implementation notes, the equivalent convex‐optimization formulation, and its solution, is provided in Appendix~\ref{apx:imputation}.

\subsection{Sessionization}
\label{section:downsampling}
\begin{figure}[h]
  \centering
  \includegraphics[width=\linewidth]{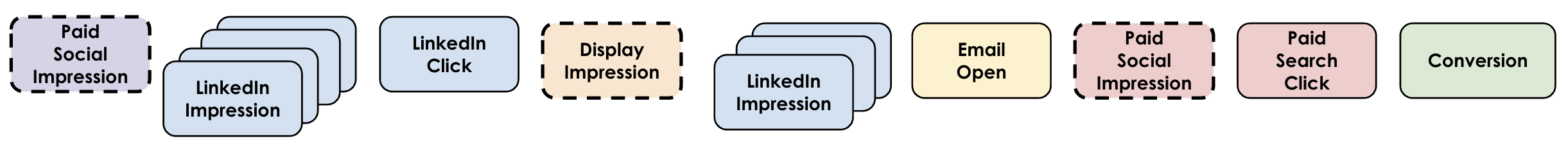}
  \caption{We group touchpoints from some owned channels, e.g., LinkedIn Platform, within a session into a single representative touchpoint. The attribution given to this one touchpoint is distributed evenly among the grouped campaigns.}
  \label{fig:downsampling}
  \Description{}
\end{figure}

Another challenge in attribution modeling is that some channels generate far more touchpoints than others, creating a signal imbalance. For example, emails may be sent weekly, whereas in-feed ads can appear multiple times per day. Since our attribution paths have a fixed length, owned-channel touchpoints can dominate and crowd out paid (external) exposures. To address this, we apply two path-adjustment steps, sessionization and downsampling, to produce more balanced engagement sequences (Figure \ref{fig:downsampling}). 

For sessionization, we group impressions from the same campaign on the same day into a single session‐level touchpoint for specific channels. For downsampling, we apply empirically derived, channel–action–specific drop rates to remove excess touchpoints. In both approaches, we recover the attribution credit to grouped/removed events according to the sampling probability. More details can be found in Appendix~\ref{apx:processing}.

\subsection{Output Calibration}
\label{section:outputScaling}
MMM provides channel-level attribution and measures marketing incrementality in the context of macroeconomic factors but cannot trace individual member journeys. DDA captures those journeys at the interaction level but lacks certain signals (e.g.\ radio, events, TV, or privacy-restricted data). Two independent attribution systems may diverge, particularly if they utilize different interactions, and thus a unified approach is needed\cite{googleMMMWhitePaper, FBRobyn}. \\
We introduce a training-time calibration that extracts signals from our MMM model to our DDA model to form a unified macro level view for marketers, while still permitting fine-grained, path-level attributions. Let $a_j(\theta)$ be the DDA attribution weight for touchpoint $j$, $A^{\rm DDA}_c(\theta)=\sum_{j:\mathrm{chan}(j)=c}a_j(\theta)$ the total DDA weight for channel $c$, $A^{\rm MMM}_c$ the corresponding channel-level attribution from MMM, $\mathcal L_{\rm DDA}(\theta)$ the usual DDA training loss, and
$\beta\ge0$ a tuning parameter. We minimize the combined loss:
\[
\mathcal L_{\rm total}(\theta)
=\mathcal L_{\rm DDA}(\theta)
\;+\;
\beta\sum_{c}\bigl(A^{\rm DDA}_c(\theta)-A^{\rm MMM}_c\bigr)^{2},
\]
and provide \textit{path level calibration} in Algorithm~\ref{algo:train_ddammm_calib}.\\
This calibration “teaches” the DDA model, during training, to respect the macro-level attribution from MMM, eliminating the need for any post-hoc rescaling, while preserving detailed, campaign- and engagement-level attribution within each channel. Details on \textit{batch level calibration} and \textit{path level calibration} can be found in Appendix~\ref{apx:batch_calib} and Appendix~\ref{apx:path_calib}.

\begin{algorithm}[h]
\caption{Train DDA Model with Path‐Level MMM Calibration}
\label{algo:train_ddammm_calib}
\begin{algorithmic}
\State \textbf{Inputs:}
\begin{itemize}
  \item Paths dataset $\{(x_i,y_i,s_i)\}_{i=1}^N$, where $x_i$ are features, $y_i$ labels, and $s_i$ the sequence of length $L_i$
  \item MMM attribution $a_c^{\rm MMM}$, channel distribution $\pi_c^{\rm MMM}$
  \item DDA network $g(\cdot;
  \theta)$ with parameters $\theta$, that output conversion probability $f$, and attention weight $a$.
  \item Hyperparameters: calibration weight $\beta$, learning rate $\eta$, epochs $T$, batch size $B$
\end{itemize}
\State \textbf{Output:} Trained parameters $\theta$
\medskip

\For{$\mathrm{epoch}=1$ to $T$}
  \State Shuffle dataset
  \For{each batch $\mathcal{B}$ of size $B$}
    \Comment{Forward pass}
    \For{each path $i\in\mathcal B$}
      \State Compute DDA attributions $a_i(c;\theta)$ for $c=1,\dots,C$
      \State Compute path-level channel distribution $\pi^{(i)}_c$
      \State Compute path‐level IPW weights $w^{(i)}_c = \frac{\pi^{(i)}_c}{\pi_c^{\rm MMM}}$
      \State "Recover" path-level MMM attribution as target
      \[
        \tilde t^{(i)}_c = w^{(i)}_c\,a_c^{\rm MMM},
        \quad
        t^{(i)}_c = \frac{\tilde t^{(i)}_c}{\sum_{d}\tilde t^{(i)}_d}
      \]
    \EndFor

    \State Compute base DDA loss
    \[
      \mathcal L_{\rm DDA}
      = \frac{1}{|\mathcal B|}\sum_{i\in\mathcal B}\ell\bigl(f_i(\cdot;\theta),y_i\bigr)
    \]
    \State Compute calibration loss (MSE or KL):
    \[
      \mathcal L_{\rm calib}
      = \frac{1}{|\mathcal B|}\sum_{i\in\mathcal B}\sum_{c=1}^C
        \bigl(a_i(c;\theta)-t^{(i)}_c\bigr)^2
    \]
    \State Total loss:
    \[
      \mathcal L = \mathcal L_{\rm DDA} \;+\;\beta\,\mathcal L_{\rm calib}
    \]
    \Comment{Backward and update}
    \State $\theta \gets \theta - \eta\,\nabla_\theta \mathcal L$
  \EndFor
\EndFor
\end{algorithmic}
\end{algorithm}

\section{Experimentation Results}

The goal of the attribution model is to proportionally quantify the impact of each marketing action towards a conversion. Since there is no ground-truth data on the contribution of each action towards a conversion, we perform the following offline and online validations to evaluate the attribution model.

\subsection{Offline Validation}

In offline validation, we seek to ensure the model’s accuracy and stability as these are intuitively desirable properties of any attribution system. We also performed ablation analysis on model input and compare our model with external models. Due to space limitation, we summarize the finding here. Further details can be found in Appendix~\ref{apx:B}.

\subsubsection{Prediction Stability}
We first assess the model's performance using binary classification metrics, aiming for high predictive accuracy. We hold 10\% of the training data for validation and separate the rest into 10 subsets. We retrain an attribution model on each subset, evaluate it on the holdout dataset, and record its performance. The resulting prediction performances converge ($>0.97$ for ROC-AUC and PR-AUC), confirming that the attribution model has high prediction power and exhibits consistent results. \\
We also evaluate the stability of predicted conversion rate using the Anderson-Darling statistic \cite{stephens1974edf, anderson1952asymptotic}. The result shows that the predicted conversion rates are consistent over time. 

\subsubsection{Attribution Stability}
\begin{figure}[h]
  \centering
  \includegraphics[width=1.0\linewidth]{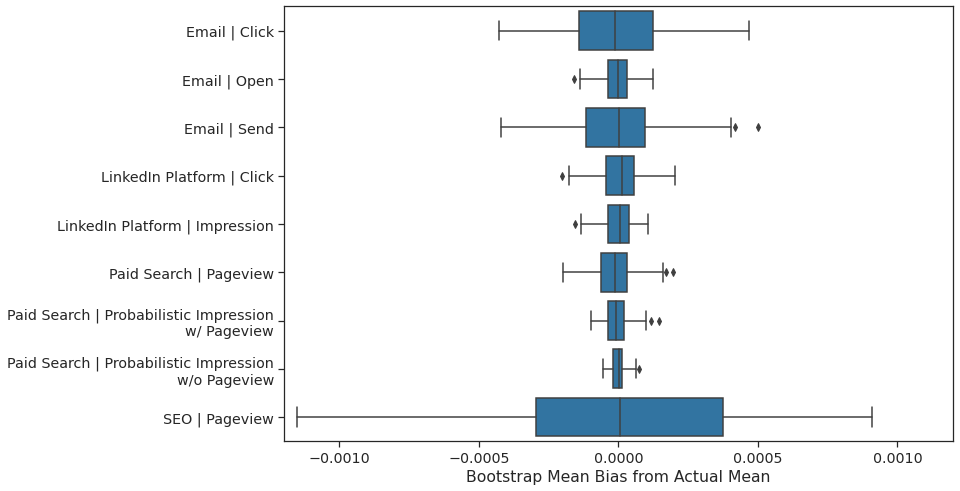}
  \caption{Distribution of bias in the bootstrap means of the action weights, calculated as the difference between each bootstrap mean and the actual sample mean. The bias distributions center around 0.}
  \label{fig:action_weight_inference}
  \Description{}
\end{figure}

We perform a similar check on our model’s stability in producing attribution results. For each action, we bootstrap sample that action’s weights with replacement for 100 times, calculate the mean of the samples, and analyze the distribution of the 100 means to evaluate the stability of the attribution weights (Figure~\ref{fig:action_weight_inference}). We observe that the mean weights of each action converge. While this approach does not have an intrinsic absolute threshold, we use these values as guardrails to monitor for potential issues.

\subsubsection{Ablation on Model Input}
We conducted an ablation study to quantify the impact of each model component by training and evaluating DDA model with particular features removed, including member/company embeddings, campaign embeddings, date embeddings, or all embeddings (sequence only). We compared ROC-AUC against the baseline and measured how the channel-level attribution mix deviated (mean squared deviation on percentage) from the full model. Removing campaign or date features caused small AUC drops (–0.1\% and –0.3\%), while omitting entity embeddings and using sequence alone reduced AUC more substantially (–1.7\% and –5.1\%). In terms of attribution stability, date removal produced the largest shift (MSD=49.3), with campaign and entity removals and the sequence-only variant showing smaller but non-negligible deviations (MSD=8.5, 14.4, and 11.8, respectively). 

\subsubsection{Approach Comparison}
We performed additional offline evaluations of LiDDA and CausalMTA \citep{yao2022causalmta} using code published by the authors on GitHub. We implemented CausalMTA as described by the authors, omitting features such as time information that are not part of its original design, and adapting inputs such as member embeddings. We trained both CausalMTA and LiDDA for 5 trials using a static set of marketing paths. In Table~\ref{tab:modelAUROCComparison} we observe the models fit equally well to the data in terms of ROC-AUC evaluation with both achieving scores between 0.982 and 0.983. 

In Table~\ref{tab:modelComparison}, we compare the calculated attribution across two marketing channels. Here we find significant differences between the models. LiDDA outperforms CausalMTA where its results are more closely aligned with our experimentally validated media-mix model which reports a mix of 89.6\% and 10.4\% across channels A and B respectively.

\begin{table}[h]
\centering
\caption{Normalized attribution by marketing channel from CausalMTA, LiDDA, and MMM ground-truth model. Results are computed over 5 trials and report 95\% CI.}
\label{tab:modelComparison}
\begin{tabular}{lccc}
\toprule
Channel & CausalMTA & LiDDA & MMM \\
\midrule
Channel A & $0.667 \pm 0.010$ & $0.826 \pm 0.001$ & 0.896 \\
Channel B & $0.333 \pm 0.011$ & $0.174 \pm 0.001$ & 0.104 \\
\bottomrule
\end{tabular}
\end{table}

Beyond attribution quality, computational efficiency is a practical consideration for production deployment. We compare the training time taken for each model on a common dataset. The CausalMTA model requires multiple training passes for the variational autoencoder, journey reweighting, domain classifier, and outcome prediction. By contrast, LiDDA is trained in a single pass and obtains weights at no additional cost from the attention layer. For this comparison we set all models to 3 epochs and trained on a common dataset with 40,000 training paths and 8,000 inference paths. 

Additionally, while the CausalMTA paper refers to Shapley values for allocation credit, the authors note that this is computationally infeasible and rely on an approximation \citep{yao2022causalmta}. The available code for CausalMTA utilizes incremental value heuristic (IVH) methodology where allocation is derived by measuring the predicted outcome change caused by each timestep. Although this method has been shown to produce inaccurate estimations \citep{singal2019}, we utilized it for our baseline since the provided Shapley calculation could not run given our path lengths. 

We made a further improvement to the provided IVH code by implementing a batched version to facilitate faster testing. This modified version was used in our benchmark. In Table~\ref{tab:modelTimingComparison}, we observe significantly faster training and inference performance from LiDDA with a 78\% reduction in training time and a 99\% reduction in inference time.

\begin{table}[h]
\centering
\caption{Mean training and inference time in seconds using all models set to 3 epochs with 95\% CI reported over 5 trials. Inference for CausalMTA uses IVH while LiDDA uses attention weights.}
\label{tab:modelTimingComparison}
\begin{tabular}{lccc}
\toprule
Stage & CausalMTA & LiDDA & Reduction \\
\midrule
Training (epoch) & $145.37 \pm 0.68$ & $31.97 \pm 0.37$ & $-78\%$ \\
Inference (dataset) & $369.72 \pm 4.12$ & $2.16 \pm 0.05$ & $-99\%$ \\
\bottomrule
\end{tabular}
\end{table}

\subsection{Online Validation of Deployment}
\label{section:onlineValidation}

Offline validation does not give us insight into whether the attention weight based attribution is close to actual member behavior. Thus, we run experiments for online validations to check how close our model’s attributions align with real-world marketing efforts. We can either run member level test for owned channels or geo-level experiments for external paid channels. 
We conduct incremental lift tests for the email channel. In particular, we randomly split members into two groups. We select certain campaigns ($\mathcal{T}^{h}$) to hold out email sending during the experiment in the control group and conduct business as usual in the treatment group. It is important to note that email sending was sparse across both members and time, limited to approximately one email per week to a small set of selected members within the connection graph. Additionally, AA tests were conducted as a basic validation of the experimental setup, and no statistically significant differences in conversion rates were found between groups.\\
To validate member-level experiment, we simulate a holdout experiment within the treatment group by creating counterfactual conversion paths, where the holdout campaign touchpoints ($t\in \mathcal{T}^h$) are omitted. This approach helps estimate what the conversion path would look like had the member been in the control group. 
Define two conversion probabilities: \(P(Y=1|Z=1, S=S_1)\) with original path $S_1$, and \(P_1(Y=1|Z=1, S=S_1-\mathcal{T}^h)\) with counterfactual path $S_1-\mathcal{T}^h$, where $Z=1$ indexes the treatment group. The corresponding attribution of touchpoints $t\in\mathcal{T}^h$ can be defined as:
\begin{align}
    a({\mathcal{T}^h}) = \frac{P_1(Y=1| S=S_1) - P_1(Y=1|S=S_1-\mathcal{T}^h)}{P_1(Y=1|S=S_1)}
    \label{eq:attr_def}
\end{align}
where, $P_1(\cdot)=P(\cdot|Z=1)$.
We compare the distribution of DDA attribution against the measured attribution from the experiment. 

A similar formulation can be constructed using $P_0(\cdot)$. We did not use this route for validation, as the simulation of counterfactual paths ($S_0+\mathcal{T}^h$) by adding touchpoints requires more assumptions such as sequencing and campaigns which may introduce bias and noise.
\subsubsection{Inverse Propensity Weighting}
From the experimentation data, we can calculate $\hat{P}_1(Y=1|S=S_1)$ and $\hat{P}_0(Y=1|S=S_1-\mathcal{T}^h)$ easily, by simply counting percentage of conversions in the treatment and control group. In order to obtain an estimate of $P_1(Y=1|S=S_1-\mathcal{T}^h)$, we use inverse propensity weighting (IPW)\cite{hernan2006estimating}. Denote $e_i$ as the propensity score for member $i$ and $Z_i$ as the indicator for member $i$ receiving campaigns as in the treatment group (=1) and for being in the control group (=0). Denote $E_{M_i}$ as the member embeddings for member $i$ that are computed by LinkedIn.

\textbf{Step 1}: Estimate each member's treatment propensity, and corresponding IPW
\begin{equation}
    e_i = P(Z_{i} = 1 | E_{M_i}), \qquad
    w_i =
    \begin{cases}
    1, & \text{if } Z_i = 1 \\
    \frac{e_{i}}{1 - e_{i}}, & \text{if } Z_i = 0
    \end{cases}
\end{equation}

\textbf{Step 2}: Compute the attribution 
\begin{align}
\hat{a}^{\exp}(\mathcal{T}^h) = \frac{\hat{P}_1(Y=1| S=S_1) - \hat{P}_1(Y=1|S=S_1-\mathcal{T}^h)}{\hat{P}_1(Y=1|S=S_1)}   
\label{eq:plug_in}
\end{align}

by plugging in $\hat{P}_1(Y=1|S=S_1)=(\sum_{i:Z_i=1}Y_i)/(\sum_{i:Z_i=1}1)$, and
\begin{align}
    \hat{P}_1(Y=1|S=S_1-\mathcal{T}^h) = \sum_{i:Z_i=0}\frac{w_i Y_i}{\sum_{i:Z_i=0}w_i}.
\end{align}
For validation, we then compute the \textit{attention based attribution}, i.e.,
\begin{align}
\hat{a}^{DDA}(\mathcal{T}^h) = \frac{\sum_{i:Z_i=1}\sum_{t\in \mathcal{T}^h} a_i^{DDA}(t)}{\sum_{i:Z_i=1} 1}
\end{align}
where $a_i^{DDA}(t)$ is attention based attribution for member $i$ for touchpoint $t$. Further details, including validation of IPW and two additional attribution measures are available in Appendix~\ref{apx:ipw}.

\subsubsection{Experiment Results}
We chose three incrementality tests in the email channel to validate the attribution model and the primary metric is attribution defined in eq. [\ref{eq:attr_def}]. Our goal is to evaluate the effectiveness of the acquisition email series. The comparisons are listed in Table~\ref{tab:validation}. 

\begin{table}
\resizebox{\columnwidth}{!}{%
  \begin{tabular}{cccl}
    \toprule
    Test index & $\hat{a}^{\exp}$ & $\hat{a}^{DDA}$ & $\hat{d}^{DDA}$ \\
    \midrule
    1 & 9.68\% (1.47\%, 20.02\%) & 7.65\% (4.95\%, 10.07\%) & -2.23\% (-5.85\%, 0.06\%) \\
    2 & 14.05\% (3.35\%, 24.75\%) & 14.96\% (11.78\%, 18.24\%) & 0.95\% (-0.89\%, 2.25\%) \\
    3 & 12.54\% (2.10\%, 27.69\%) & 15.21\% (11.65\%, 19.34\%) & 2.41\% (-0.38\%, 5.05\%) \\
    \bottomrule
  \end{tabular}
  }
  \caption{Experimental validation for 3 incrementality tests. Each cell contains the point estimate and its corresponding 95\% confidence intervals.}
  \label{tab:validation}
\end{table}

We compare DDA generated attribution $\hat{a}^{DDA}$ with "ground truth" $\hat{a}^{\exp}$ and compute their difference, i.e. $\hat{d}^{DDA} = \hat{a}^{DDA} - \hat{a}^{\exp}$.

Confidence intervals of $\hat{a}^{\exp}$, $\hat{a}^{DDA}$, and $\hat{d}^{DDA}$ were derived using bootstrap sampling. We notice that the gaps between $\hat{a}^{\exp}$ and $\hat{a}^{DDA}$ are not statistically significant. We concluded that our DDA attention based attribution aligns with the experimental result.


\section{Application to LinkedIn Ads Measurement}
The LiDDA model architecture was first applied to LinkedIn’s GTM marketing. We have since integrated a variant of LiDDA into LinkedIn’s Ad Platform to provide advertisers with comprehensive funnel insights through LinkedIn Ads. Previously, measurement of ad impact on the Ads platform heavily relied on last-touch attribution, which often undervalues upper-funnel ads, potentially resulting in suboptimal budget allocation. In contrast, data-driven attribution distributes credit across multiple ad interactions throughout the buyer’s journey, accounting for the impact of upper-funnel advertising efforts on business outcomes.

\begin{figure}[h]
    \centering
    \includegraphics[width=0.75\linewidth]{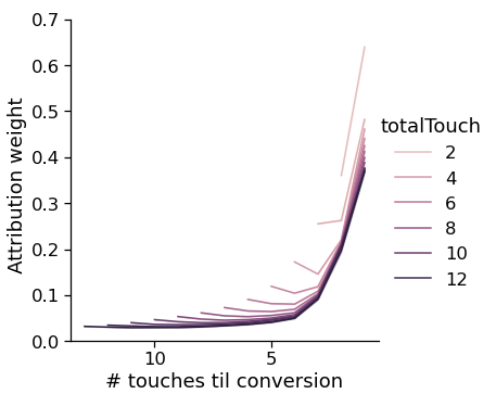}
    \caption{Attribution weights for LinkedIn lead conversions show a time decay effect in aggregate; ad touches closer to the conversion tend to get more credit.
}
    \label{fig:lms_attribution_weights_agg}
\end{figure}

When applying LiDDA to the LinkedIn Ads use case, we also supply information about the advertiser, via a company embedding for the advertiser. This is concatenated with embeddings representing the member and their company. Similar to our GTM marketing model, we also include a touchpoint sequence representation (encoding the sequence of touch types, e.g., "Onsite lead generation sponsored status update video ad impression"), touchpoint embeddings derived from campaign text, and positional encodings (representing the time between each touchpoint and the conversion).
While GTM marketing has access to touchpoints across multiple channels, the Ads platform currently only has access to LinkedIn touchpoints. When applying the model to lead conversions obtained through LinkedIn’s Lead Generation form, on average the attribution weights show a time decay effect: touches closer to the conversion tend to get more credit (Figure~\ref{fig:lms_attribution_weights_agg}). Meanwhile, we also observe the weight assigned to a specific touchpoint can vary, influenced by factors like the objective type, ad format, interaction type (impression, click), ad creative, days before conversion. We surface these insights as both path‐level detail and aggregated campaign/creative reports (Details in Appendix~\ref{apx:ad_platform}).

\section{Conclusion}
In this paper, we introduce LiDDA, a unified transformer-based data-driven attribution approach we developed and deployed at LinkedIn. We demonstrated the innovations to DDA methodology that combine an attention mechanism, positional embeddings, touchpoint imputation, and MMM calibration. We also shared learnings from implementation and deployment that are broadly applicable to the marketing and ad tech fields.

\begin{acks}
We thank Licurgo Benemann De Almeida, Artem Grigoryan, GC Chapiewski, Sabry Tozin, Shruti Sharma, Zaid Nasir, Anu Bedi, Paolo Provinciali for support of the work. 
\end{acks}

\bibliographystyle{ACM-Reference-Format}
\bibliography{mta}

\appendix

\section{Additional Details on Methods}
\label{apx:A}
\subsection{Three‐Step Member Impression Allocation}
\label{apx:imputation}
Let \(d\) index days, \(c\) index paid channels, \(j\in J_{i,c,d}\) index click‐events for member \(i\), and \(k\in\mathcal P_d\) index no‐click paths on day \(d\).  We observe external aggregates \((I^{\rm ext}_{c,d},\,C^{\rm ext}_{c,d})\), owned‐channel totals \((I^{\rm own}_{\rm total},C^{\rm own}_{\rm total},I^{\rm own}_{\rm clickers},C^{\rm own}_{\rm clickers})\), and a nonnegative weight \(u_k\) for each no‐click path.

\subsubsection{Member‐Only Impression Budget}  
Compute the number of impressions on channel \(c\) and day \(d\) attributable to known members by scaling the external impressions by the fraction of clicks from members:
\[
C^{\rm mem}_{c,d}
=\sum_{i}\lvert J_{i,c,d}\rvert,
\qquad
I^{\rm mem}_{c,d}
=I^{\rm ext}_{c,d}\,\frac{C^{\rm mem}_{c,d}}{C^{\rm ext}_{c,d}}.
\]
Summing over channels gives the total daily member‐impression budget:
\[
I_d \;=\;\sum_{c}I^{\rm mem}_{c,d}.
\]

\subsubsection{Impute and Pre‐Allocate to Clicks}  
First, derive a down‐weight factor \(\alpha<1\) from owned‐channel behavior:
\[
R^{\rm own}_{\rm all}
=\frac{I^{\rm own}_{\rm total}}{C^{\rm own}_{\rm total}},
\quad
R^{\rm own}_{\rm clickers}
=\frac{I^{\rm own}_{\rm clickers}}{C^{\rm own}_{\rm clickers}},
\quad
\alpha
=\frac{R^{\rm own}_{\rm clickers}}{R^{\rm own}_{\rm all}}.
\]
Apply \(\alpha\) to the raw external impressions‐per‐click to obtain the clickers’ expected impressions per click on \((c,d)\):
\[
R^{\rm ext}_{\rm all}
=\frac{I^{\rm ext}_{c,d}}{C^{\rm ext}_{c,d}},
\qquad
r_{c,d}
=\alpha\,R^{\rm ext}_{\rm all}
\;\approx\;\mathbb{E}\bigl[\#\mathrm{imps}\mid\mathrm{click},c,d\bigr].
\]
Then allocate to each click \(j\in J_{i,c,d}\) at least one impression by either
\[
m_j
=\max\bigl(1,\lfloor r_{c,d}\rfloor\bigr)
\quad\text{(deterministic)},\] 

\[m_j\sim\mathrm{Geom}\bigl(p=1/r_{c,d}\bigr)
\quad\text{(stochastic)},
\]
and compute the pre‐allocated sum and leftover:
\[
I^{\rm pre}_d=\sum_j m_j,
\qquad
L_d=I_d-I^{\rm pre}_d.
\]

\subsubsection{Probabilistic Allocation of Leftover}  
Assign the remaining \(L_d\) impressions across no‐click paths \(k\in\mathcal P_d\) proportionally to weights \(u_k\).  Define
\[
p_k=\frac{u_k}{\sum_{\ell\in\mathcal P_d}u_\ell},
\qquad
\lambda_k=L_d\,p_k.
\]
Draw
\(\tilde I_k\sim\mathrm{Poisson}(\lambda_k)\)
independently, then rescale to exactly exhaust \(L_d\):
\[
I_k
=\Bigl\lfloor \tilde I_k \times \frac{L_d}{\sum_{\ell}\tilde I_\ell}\Bigr\rfloor,
\qquad
\sum_{k}I_k=L_d.
\]

\subsubsection{Result}  
Each click‐event \(j\) receives \(m_j\ge1\) (on average \(r_{c,d}\)), each no‐click path \(k\) receives \(I_k\ge0\), and
\[
\sum_j m_j \;+\;\sum_k I_k \;=\; I_d,
\]
so the member‐impression budget is exactly allocated.

\subsubsection{Remarks} We note that:
\begin{itemize}
    \item Both Geometric distribution and Poisson distribution can be approximated by Normal distribution if needed; 
    \item $u_k \propto \mathbb{E}\bigl[\#\mathrm{imps}\mid\mathrm{noClick},k\bigr]$ can be assigned by contextual information such as corresponding campaign audience and location;
    \item $r_{c,d}$ can be estimated with more complex methods, e.g., $r_{i,c,d} = f(x_i, x_c, x_d)$ via a NN that leverages both member level data from owned channels and aggregate data from external channels. 
\end{itemize}

\subsubsection{Joint Convex‐Optimization Formulation}

Instead of treating click allocation (Step 2) and leftover allocation (Step 3) separately, we can solve them in one convex‐optimization step.

Introduce: (i) Variables \(x_j\) for the impressions assigned to click‐event \(j\in J_{*,*,d}\); (ii)  Variables \(y_k\) for the impressions assigned to no‐click path \(k\in\mathcal P_d\); (iii) Total budget \(I_d\) from Step 1;  (iv) Positive weights
  \[
    w_j \;=\; r_{c(j),d}
    \quad\bigl(\approx\mathbb{E}[m_j]\bigr),
    \qquad
    u_k \;\ge\;0
  \]
reflecting the expected imps‐per‐click and the no‐click path priorities.\\
We then solve the following optimization:

\[
\begin{aligned}
  &\underset{\{x_j\},\{y_k\}}{\text{maximize}}
  && 
    \sum_{j\in J_{*,*,d}} w_j\,\log(x_j)
    \;+\;\sum_{k\in\mathcal P_d} u_k\,\log(y_k)
  \\[6pt]
  &\text{subject to}
  && 
    \sum_{j\in J_{*,*,d}}x_j \;+\;\sum_{k\in\mathcal P_d}y_k
      \;=\;I_d,
    \\[-2pt]
  &&&
    x_j \;\ge\;1,
    \quad
    y_k \;\ge\;0.
\end{aligned}
\]
\textit{KKT‐based solution.}  The Karush–Kuhn–Tucker conditions imply that at optimum there is a multiplier \(\lambda>0\) such that

\[
x_j = \max\!\Bigl(1,\;\frac{w_j}{\lambda}\Bigr),
\qquad
y_k = \frac{u_k}{\lambda},
\]
and $\lambda$ must satisfy:
\begin{align}
    \sum_j \max\!\Bigl(1,\;\frac{w_j}{\lambda}\Bigr)
   +\sum_k \frac{u_k}{\lambda}
   = I_d. \label{eq:lambda}
\end{align}
Eq.~\ref{eq:lambda} can be solved by Algorithm~\ref{algo:compute_lambda}, with complexity $O(J\log(J))$.\\
To summarize, the solution:\\
1. Honors the total budget: \(\sum_jx_j+\sum_ky_k=I_d\). \\ 
2. Guarantees each click‐bucket \(x_j\ge1\) whenever \(w_j>0\).  \\
3. Allocates impressions in proportion to the expected click demand \(w_j\) and the no‐click weights \(u_k\).\\
By choosing \(w_j=r_{c,d}\) and \(u_k\) as in Steps 2–3, we recover a deterministic analogue of the previous stochastic Geometric+Poisson draws.

\begin{algorithm}[h]
\caption{Compute Optimal \(\lambda\) and Allocations}
\label{algo:compute_lambda}
\begin{algorithmic}
\State \textbf{Input:} Click‐weights \(\{w_j\}_{j=1}^J\), path‐weights \(\{u_k\}_{k=1}^K\), total budget \(I_d\)
\State \textbf{Output:} Impressions \(\{x_j\}, \{y_k\}\)

\State Sort \(w_{(1)} \ge w_{(2)} \ge \dots \ge w_{(J)}\)
\State Compute \(S_u \gets \sum_{k=1}^K u_k\)
\State \(S_w \gets \sum_{j=1}^J w_{(j)}\)

\For{\(m = 0\) \textbf{to} \(J\)}
  \State \(S_w^{(m)} \gets S_w - \sum_{j=1}^m w_{(j)}\)
  \State \(\displaystyle \lambda_m \gets \frac{S_w^{(m)} + S_u}{\,I_d - m\,}\)
\EndFor

\State Find the unique \(m^*\) such that
\[
  w_{(m^*)} \;\ge\;\lambda_{m^*}
  \quad\text{and}\quad
  w_{(m^*+1)} \;<\;\lambda_{m^*},
\]
with conventions \(w_{(0)}=\infty\), \(w_{(J+1)}=0\).
\State \(\lambda \gets \lambda_{m^*}\)

\For{each click \(j=1,\dots,J\)}
  \State \(x_j \gets \max\bigl(1,\;w_j/\lambda\bigr)\)
\EndFor

\For{each path \(k=1,\dots,K\)}
  \State \(y_k \gets u_k/\lambda\)
\EndFor

\end{algorithmic}
\end{algorithm}

\begin{figure}[h]
  \centering
  \includegraphics[width=\linewidth]{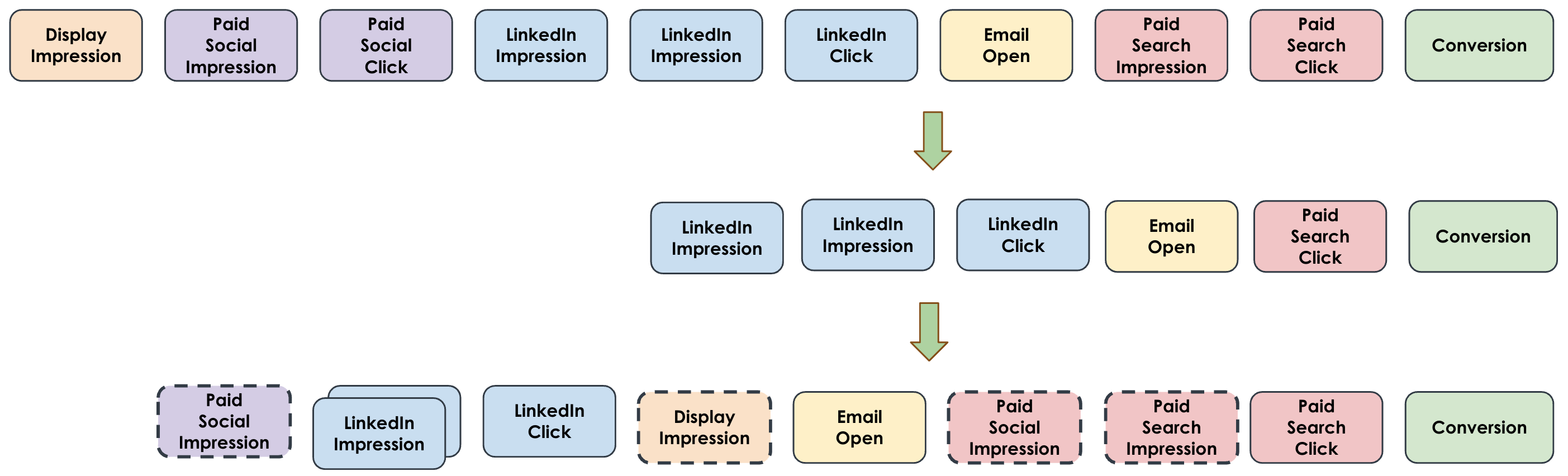}
  \caption{The top row shows an original path with all impressions and clicks a member has interacted with prior to making a conversion. We cannot collect any paid media impressions due to privacy regulations like GDPR and CCPA, and we use a fixed length for paths to use in modeling. The middle row shows the same path after these interactions are dropped. We impute probabilistic impressions into the paths and group impressions from owned channels, LinkedIn Platform as shown in the bottom row for modeling.}
  \Description{}
  \label{fig:prob-paths}
\end{figure}

\begin{figure}[H]
  \centering
  \includegraphics[width=\linewidth]{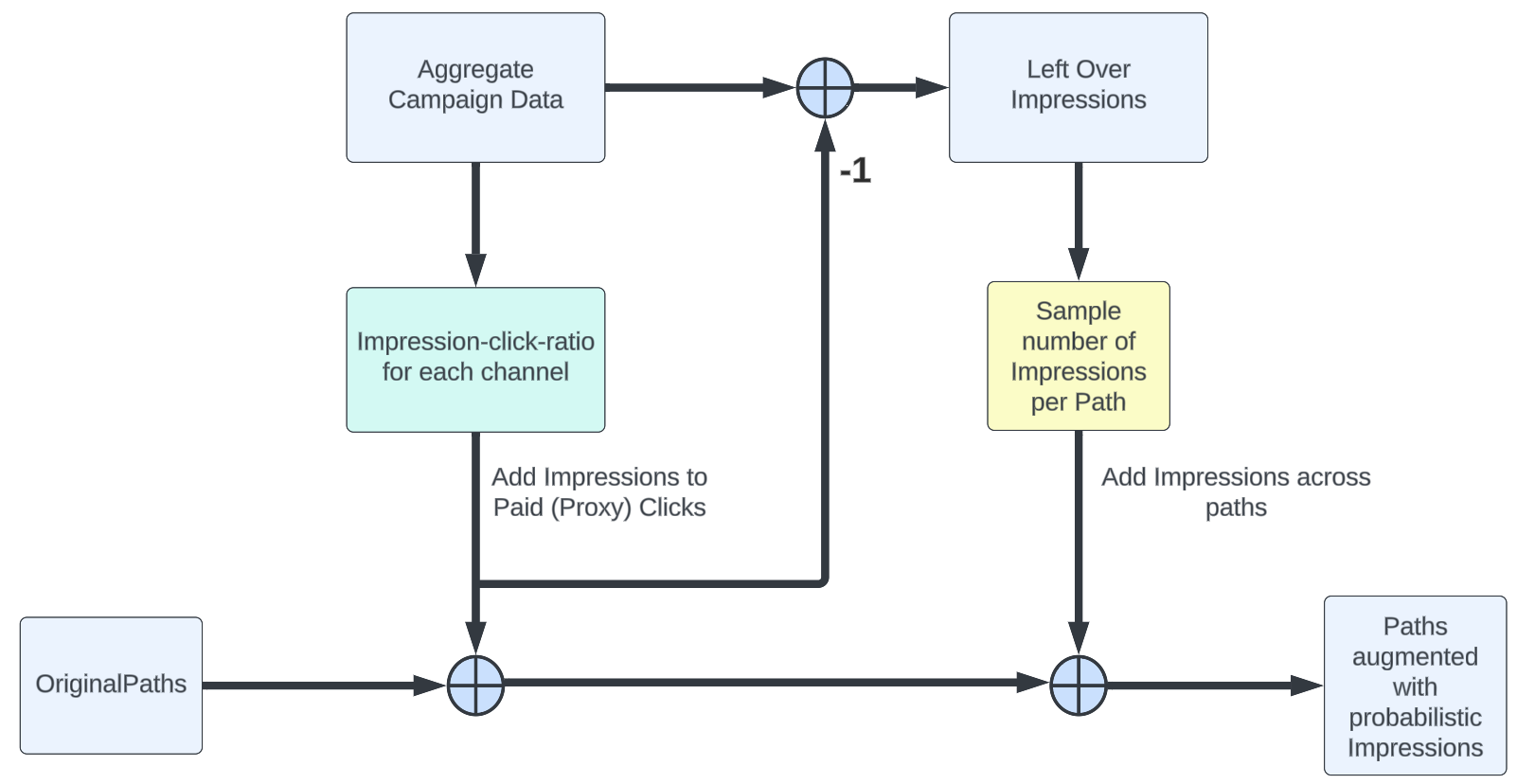}
  \caption{Probabilistic touchpoints are imputed in paths in two steps. First, we add impressions to existing proxy-click events based on impression to click ratio of each channel. Second, we distribute the remaining impressions across members online each day.}
  \Description{}
  \label{fig:prob-imps}
\end{figure}

\subsection{Path Pre-processing}
\label{apx:processing}
\subsubsection{Sessionization}
We group impressions from certain channels in a path as a single touchpoint within a session if they are from the same campaign on the same day, as detailed in Algorithm \ref{alg:sessionization}. This helps reduce noise, particularly from on-platform channels where a member may observe the same campaign repeatedly due to scrolling behavior. With this approach, we model a single event; however, we retain the downsampling links for post-model processing to reallocate credit equally across the grouped events.

\subsubsection{Downsampling}
We further address signal bias by adjusting the retention rate of specific channel–action combinations according to tunable downsampling rates. We obtain these rates using an independent analysis to estimate the comparable impact of each channel using signals such as spend and outcomes. With these ratios, we randomly "drop" some touchpoints to yield the expected ratio as described in Algorithm \ref{alg:downsampling}. We observe that this pre-processing step yields more reliable attribution results and path compositions better aligned with aggregate data.

Our attribution system is complex and heterogeneous, as a result, certain heuristics are necessary. Meanwhile, we perform comprehensive evaluation on these heuristics. For example, to downsample owned channel activities, we estimate a scaling factor to bring the number of actions, clicks/impressions, to the same order of magnitude of actions from other channels.  And we evaluate the impact through offline and online experiments. 


\begin{algorithm}
\caption{Sessionization methodology}\label{alg:sessionization}
\begin{algorithmic}
\Require Member $path$ with marketing $touchpoint$ metadata
\For{$path$}
    \State Group all impressions by $campaign$, $channel$, and $date$
    \For{$group$}
        \State Select most recent $touchpoint$ by timestamp
        \State Record array of remaining $touchpoint$
        \State Store array in $touchpoint$ metadata
    \EndFor
    \State Output truncated $path$ with linkage metadata
\EndFor
\end{algorithmic}
\end{algorithm}

\begin{algorithm}
\caption{Downsampling methodology}\label{alg:downsampling}
\begin{algorithmic}
\Require Member $path$ with marketing $touchpoint$ metadata
\Require Sampling $ratio$ per $action$ and $channel$ pair
\For{$path$}
    \State Group all touchpoints by $action$ and $channel$
    \For{$group$}
        \State Sort $touchpoints$ by timestamp
        \State Compute number of $touchpoints$ \(N\) based on $ratio$
        \State Retain \(N\) latest $touchpoints$
    \EndFor
    \State Output truncated $path$
\EndFor
\end{algorithmic}
\end{algorithm}

\subsection{Training Time Calibration}
\label{apx:trCalibration}
\subsubsection{Batch‐Level Calibration}\label{apx:batch_calib}

Let a training \emph{batch} consist of \(N\) paths.  Denote by
\[
a^{\rm DDA}_i(c;\theta)=\sum_{t\in c} a_i^{\text{DDA}}(t;\theta), 
\quad(i=1,\dots,N,\;c=1,\dots,C)
\]
the DDA model’s attribution share for channel \(c\) on path \(i\).  Define the batch‐aggregated channel totals
\[
A^{\rm DDA}_c(\theta)
=\frac{1}{N}\sum_{i=1}^N a^{\rm DDA}_i(c;\theta),
\qquad
A^{\rm MMM}_c
= a^{\rm MMM}_c
\]
where \(a^{\rm MMM}_c\) is the global MMM‐derived share for channel \(c\).  We then add one of the following \emph{batch‐level} penalties:\\
\textit{Mean‐Squared Error (MSE) Penalty}
\[
\mathcal L_{\rm calib}^{\rm batch}
=\sum_{c=1}^C 
   \Bigl(A^{\rm DDA}_c(\theta) - A^{\rm MMM}_c\Bigr)^2;
\]
\textit{Kullback–Leibler (KL) Penalty}
\[
\mathcal L_{\rm calib}^{\rm batch}
=\sum_{c=1}^C 
   A^{\rm MMM}_c \,\ln\!\frac{A^{\rm MMM}_c}{A^{\rm DDA}_c(\theta)}.
\]

These terms are computed once per batch and combined with the base DDA loss:
\[
\mathcal L_{\rm total}
=\mathcal L_{\rm DDA}(\theta)
\;+\;\beta\,\mathcal L_{\rm calib}^{\rm batch}.
\]
Backpropagation propagates gradients from the channel‐sum residuals into each \(a^{\rm DDA}_i(c;\theta)\).
\bigskip
\subsubsection{Path‐Level Calibration}\label{apx:path_calib}

For each path \(i\), let \(\pi^{(i)}_c\) be the empirical fraction of touches in channel \(c\):
\[
\pi^{(i)}_c
=\frac{\#\{\text{touchpoints of }c\text{ in path }i\}}{\text{length}(i)}.
\]
Let \(\pi^{\rm MMM}_c\) be the probability of any touchpoint's channel being $c$ (i.e., estimate from all paths), and $a_c^{MMM}$ be the "attribution" of $c$ from MMM.  Define a per‐path, per‐channel weight
\[
w^{(i)}_c
=\frac{\pi^{(i)}_c}{\pi^{\rm MMM}_c}.
\]
Form the unnormalized target
\[
\tilde t^{(i)}_c
= w^{(i)}_c \;a^{\rm MMM}_c,
\]
and then normalize to "recover" a valid path-level MMM attribution
\[
t^{(i)}_c
=\frac{\tilde t^{(i)}_c}{\sum_{d=1}^C \tilde t^{(i)}_d},
\quad
\sum_{c=1}^C t^{(i)}_c = 1.
\]

We then penalize the divergence between DDA’s prediction and this target for each path:\\
\textit{Per‐Path MSE}
\[
\mathcal L_{\rm calib}^{\rm path}
=
 \sum_{c=1}^C
 \Bigl(
   a^{\rm DDA}_i(c;\theta) - t^{(i)}_c
 \Bigr)^2.
\]
\textit{Per‐Path KL}
\[
\mathcal L_{\rm calib}^{\rm path}
=
 \sum_{c=1}^C
   t^{(i)}_c \,\ln\!\frac{t^{(i)}_c}{a^{\rm DDA}_i(c;\theta)}.
\]
As before, the total loss is
\[
\mathcal L_{\rm total}
=\mathcal L_{\rm DDA}(\theta)
\;+\;\beta\,\mathcal L_{\rm calib}^{\rm path},
\]
and gradients flow from each path‐level residual into the DDA model parameters \(\theta\).\\ 

\textit{Remarks on implementation}:
\begin{enumerate}
    \item When there are padding in the path, $\pi_c^{MMM}$ needs to take into account the padding, e.g., $\pi_c^{MMM} = \frac{1}{\mathcal{N}}\sum_i \pi_c^{(i)}$ if computation of $\pi_c^{(i)}$ throws away the padding. Here, $\mathcal{N}$ denotes the number of paths in the full dataset. 
    \item Since we applied scaling on the (theoretical IPW) target $\tilde t$, we can apply the weight back to the loss function when using the normalized target $t$, i.e., $\mathcal L_{\rm calib}^{\rm path}
=
 \sum_{c=1}^C
 v^{(i)}\Bigl(
   a^{\rm DDA}_i(c;\theta) - t^{(i)}_c
 \Bigr)^2.$
 where, $v^{(i)} = \sum_{d=1}^C \tilde t^{(i)}_d$.
\end{enumerate}

\section{Additional Offline Analysis}
\label{apx:B}
\subsection{Model Stability}
\label{apx:stablility}
\begin{figure}[h]
    \centering
    \includegraphics[width=1\linewidth]{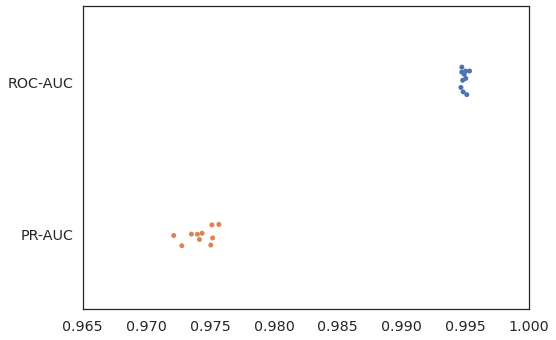}
    \caption{The ROC-AUC and PR-AUC of the attributions models retrained on 10 different subsets of training data and evaluated on the same holdout dataset. The models' performances converge.
}
    \label{fig:auc_stability}
\end{figure}

\begin{figure}[h]
  \centering
  \includegraphics[width=\linewidth]{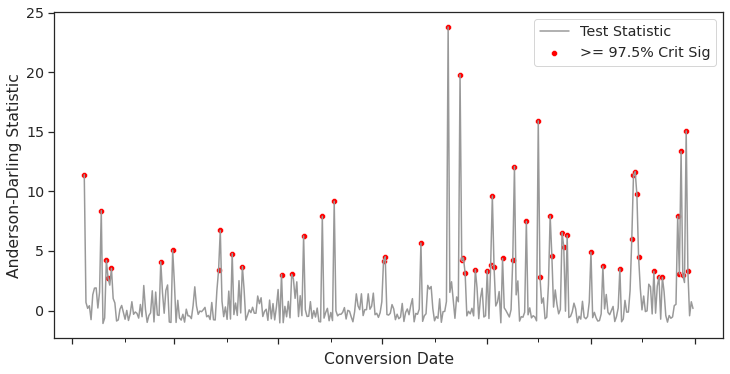}
  \caption{Anderson-Darling statistic comparing the distribution of conversion propensities each day to the distribution from the day a week ago. Days where the propensity distributions are significantly different are marked in red.}
  \label{fig:andersonDarling}
  \Description{}
\end{figure}

We provide visualizations of model stability on prediction and attribution in Figure~\ref{fig:auc_stability}, Figure~\ref{fig:andersonDarling} and Figure~\ref{fig:action_weight_inference}.

\subsection{Attention Interpretability}
\label{subsection:interpretability}
When using attention weights in attribution, we would expect that the computed weights quantify the contribution a touchpoint has towards the prediction of a conversion outcome. Therefore, if we were to modify the weight distribution we should observe a change in the predicted outcome. In this test, we train two models on the same dataset. The baseline model is trained using the standard attention setup described in this paper. The modified model randomly permutes the attention weights at a path-level during inference forward passes.

We train both models for 10 epochs each using identical batch sizes and hyperparameters. After training, we perform an inference pass on a shared holdout dataset and measure ROC-AUC. For the baseline model, we perform a single inference pass. For the modified model, we perform 10 inference passes to capture the variance introduced by the weight permutation process. We repeat this entire process 30 times for each model.

\begin{figure}
    \centering
    \includegraphics[width=1\linewidth]{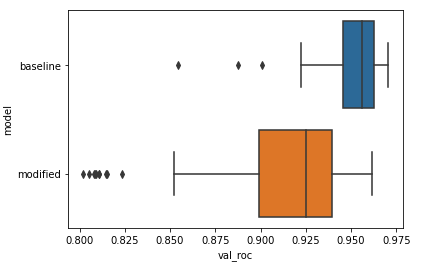}
    \caption{The plot compares the ROC-AUC against a common holdout for both the baseline and modified attention models in our study. The baseline model consists of 30 datapoints with 1 validation step per trial. The modified model consists of 300 datapoints with 10 validation steps per trial.}
    \label{fig:a1_overall}
\end{figure}

In Figure~\ref{fig:a1_overall}, we observe that the baseline model generally yields higher performance with a median ROC-AUC of 0.956 while the modified model has a median ROC-AUC value of 0.928 (-2.93\%). While this is not a substantial change, we observe a larger negative impact on the consistency of results. The AUCs of both models are noted to be high. There are two explanations for this. One reason is due to resampling of the data creating an artificially high rate of conversions. The model is measured on uncalibrated results. Secondly, we note that in our domain, converting paths have a notably different composition which facilitates an easier classification task.

\begin{figure}
    \centering
    \includegraphics[width=1\linewidth]{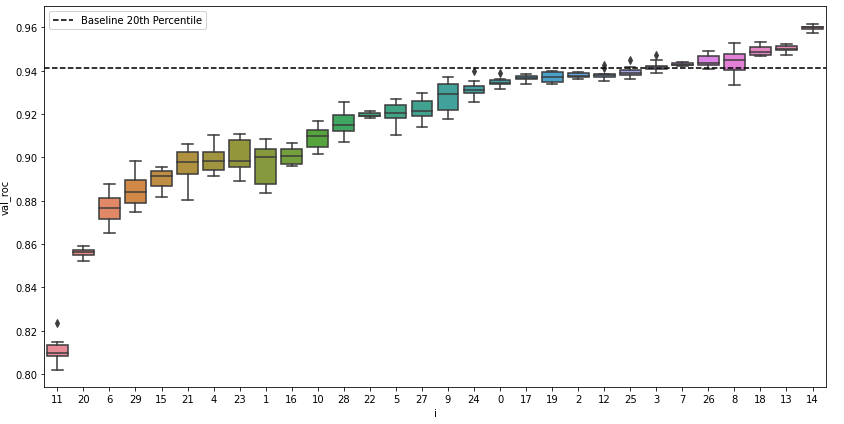}
    \caption{This plot shows the per-trial results for the modified model. The trials are sorted in ascending order according to their median ROC-AUC value. We observe that 23 of the 30 models fall below the bottom 20th percentile of the baseline trials.}
    \label{fig:a1_detail}
\end{figure}

In Figure~\ref{fig:a1_detail}, we plot the results of the 30 modified trials sorted in median ascending order. 23 of the 30 trials ranked below the 20th percentile compared to the baseline model. We conclude from this test that the introduction of weight noise has a material impact on model performance and therefore provides some explainability power.

\subsection{Ablation}
\label{apx:ablation}
\subsubsection{Methodology}
We performed an ablation study to estimate the contribution of different components of our model introduced in this paper. We generated fixed training, validation, and inference datasets to use across each simulation where a new model was trained and inferenced with the indicated component removed. In each simulation, the removal of the component was performed by modifying the code to omit the component from the graph. All hyperparameters were held constant, and each model was trained for 5 epochs. 

After training, we evaluated the ROC-AUC metrics against the validation dataset which contains a resampled population ratio of non-converting to converting of 10:1. We also evaluated the attribution weight distribution across channels for each of the models.

We tested the following components of the model:
\begin{itemize}
    \item \textbf{Baseline:} The unmodified model as described in the paper
    \item \textbf{No-Entity:} Member and company embedding inputs removed
    \item \textbf{No-Campaign:} Touchpoint campaign embedding inputs removed
    \item \textbf{No-Date:} Touchpoint date embedding inputs removed
    \item \textbf{Seq-Only:} All embeddings removed leaving only the sequence and positional encoding
\end{itemize}

\subsubsection{Classification Results}
We first evaluated the ablation effects on the final classification power of the model by comparing the predicted model score to the ground-truth label. We note high nominal AUC values as the simulations were performed on a resampled dataset. We observe that there is a moderate performance loss for the touchpoint level date and campaign features. Removal of entity embeddings had a larger effect, and removal of all components had the most negative effect. We also note that utilizing the sequence only allows for substantial predictive power given that converting members in our domain behave differently than non-converting members.

\begin{table}[h]
    \centering
    \begin{tabular}{ccc}
        \toprule
        Version & ROC-AUC & \% Change \\

        \midrule
        1. Baseline & 0.990 &  --\\
        2. No-Campaign & 0.989 & -0.1\% \\
        3. No-Date & 0.984 & -0.3\% \\
        4. No-Entity & 0.973 & -1.7\% \\
        5. Seq-Only & 0.939 & -5.1\% \\
        \bottomrule
    \end{tabular}
    \caption{ROC-AUC performance measured on a constant validation dataset after 5 epochs of training. Percentage change vs. the baseline model is shown.}
    \label{tab:ablation-auc}
\end{table}

\subsubsection{Weight Results}
We next evaluated the impact of the component removal on the measured attribution credit across a grouping of our four marketing channel families. We compute the mean squared deviation in total attribution from the baseline model across the four channels as a measure of volatility in the ablation. We observe that all ablations result in some changes to attribution however removing the dates alone causes the most significant shifts.

\begin{table}[h]
    \centering
    \begin{tabular}{cc}
        \toprule
        Version & Mean Squared Deviation \\
        \midrule
        1. Baseline & -- \\
        2. No-Campaign & 8.5 \\
        3. No-Date & 49.3 \\
        4. No-Entity & 14.4 \\
        5. Seq-Only & 11.8 \\
        \bottomrule
    \end{tabular}
    \caption{The percent attribution allocated across four marketing channels is computed using the indicated LiDDA model version. The absolute deviation for each channel vs. the baseline model is computed. The reported number is the mean squared deviation for that model, across all channels, compared with the original baseline mix.}
    \label{tab:ablation-wt}
\end{table}

\subsection{Model ROC-AUC Comparison}
Additional comparison of ROC-AUC performance between CausalMTA and LiDDA on common training dataset.

\begin{table}[h]
\centering
\caption{ROC-AUC score (95\% CI) comparison over 5 trials using static training and validation datasets.}
\label{tab:modelAUROCComparison}
\begin{tabular}{lc}
\toprule
Model & ROC-AUC Score \\
\midrule
CausalMTA & $0.983 \pm 0.001$ \\
LiDDA & $0.982 \pm 0.000$ \\
\bottomrule
\end{tabular}
\end{table}

\subsection{Recovering Distribution}
\label{apx:ipw}
\begin{figure}[h]
  \centering
  \includegraphics[width=0.8\linewidth]{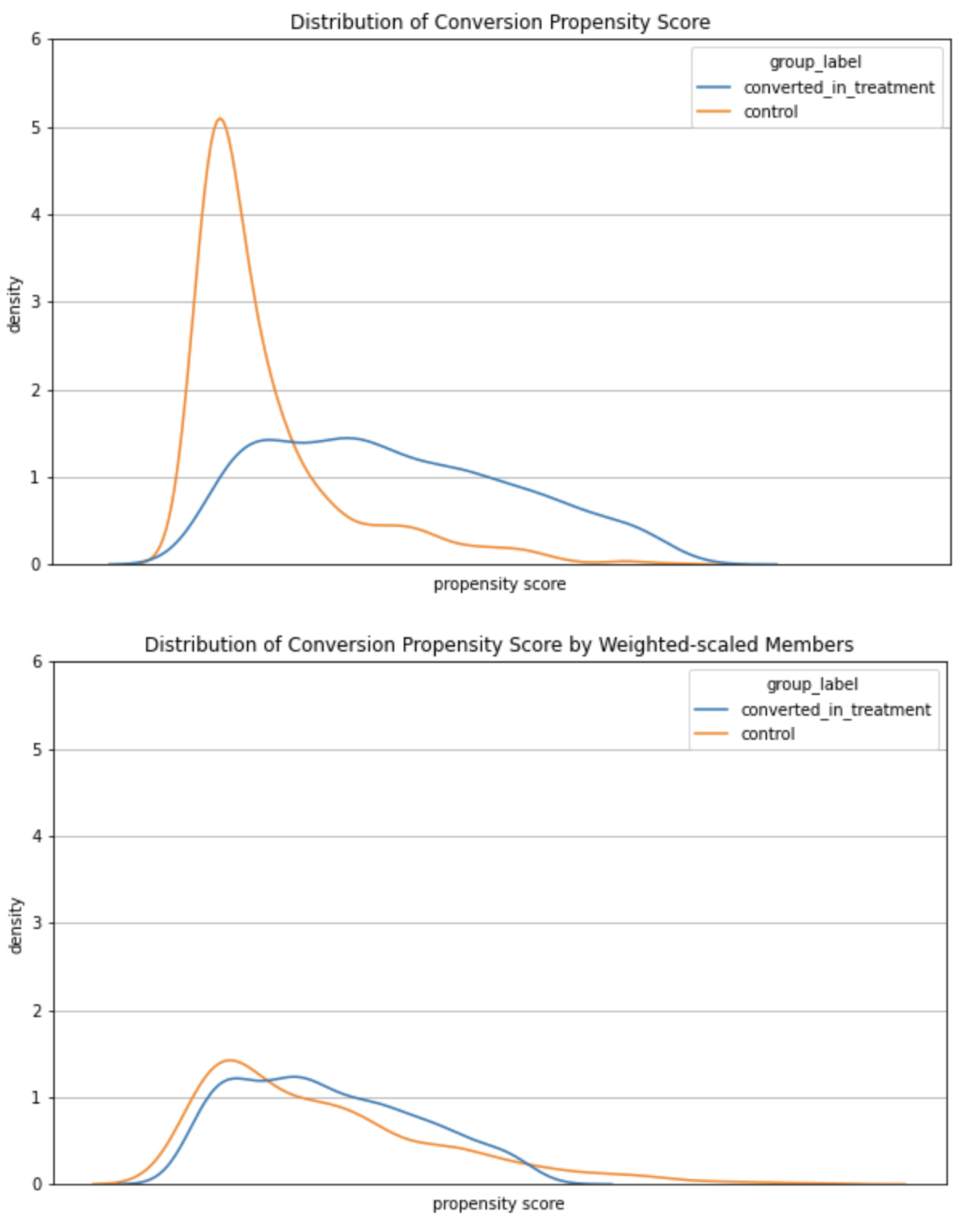}
  \caption{The top chart shows a mismatch between conversion propensity score distribution for converted members in the treatment group vs. control members. The bottom chart shows that conversion propensity score distributions are better matched in the pseudo-populations where we replicate the members through $w_i$ to balance the two populations.}
  \label{fig:ipw}
  \Description{}
\end{figure}

The inverse propensity score weight can help balance the converted members in the treatment group and members in control group for fairer comparison between model predicted conversions and observed conversions from online experiments. In Figure ~\ref{fig:ipw}, we create weight-scaled pseudo-populations through $w_i$ for those who converted in the treatment group and for those in the control group. We observe that IPW functions as expected to adjust the population from converted members in the treatment group and control group to a closer match. 

Without IPW, we can calculate a raw estimate of $a^{\text{raw}}({\mathcal{T}^h})$, which can be biased. 
\begin{align*}
    a^{\text{raw}}({\mathcal{T}^h}) = \frac{\hat{P}_1(Y=1| S=S_1) - \hat{P}_0(Y=1|S=S_1-\mathcal{T}^h)}{\hat{P}_1(Y=1|S=S_1)}.
\end{align*}
\begin{table}
 \resizebox{\columnwidth}{!}{%
  \begin{tabular}{cccl}
    \toprule
    Test index & $a^{\text{raw}}$ & $\hat{a}^{DDA}$ & $\tilde{a}^{DDA}$ \\
    \midrule
    1 & 8.72\% (0.01\%, 16.11\%) & 7.65\% (4.95\%, 10.07\%) & 12.54\% (10.52\%, 14.73\%) \\
    2 & 9.73\% (1.71\%, 17.75\%) & 14.96\% (11.78\%, 18.24\%) & 19.61\% (16.81\%, 23.44\%)\\
    3 & 16.65\% (3.93\%, 29.37\%) & 15.21\% (11.65\%, 19.34\%) & 9.54\% (7.53\%, 11.75\%)\\
    \bottomrule
  \end{tabular}%
  }
  \caption{Additional comparison for 3 incrementality tests. Each cell contains estimated attribution and its corresponding 95\% confidence intervals.}
  \label{tab:validation_ipw}
\end{table}
We also calculate a plug-in estimate $\tilde{a}^{DDA}$ of the attribution with DDA model, similar to eq. [\ref{eq:plug_in}]. Let $\pi(S)$ be the predicted conversion for touchpoint sequence $S$ from DDA model,  we can compute $\tilde{P}_1(Y=1|S=S_1)=(\sum_{i:Z_i=1}\pi(S_{1,i}))/(\sum_{i:Z_i=1}1)$, and $\tilde{P}_1(Y=1|S=S_1-\mathcal{T}^h) = (\sum_{i:Z_i=1} \pi(S_{1,i}-\mathcal{T}^h))/(\sum_{i:Z_i=1}1)$. We compared these two estimates with "ground truth" $\hat{a}^{\exp}$ in Table~\ref{tab:validation_ipw}. 
\subsection{Application to Ad Platform}
\label{apx:ad_platform}

\begin{figure}[h]
    \centering
    \includegraphics[width=0.75\linewidth]{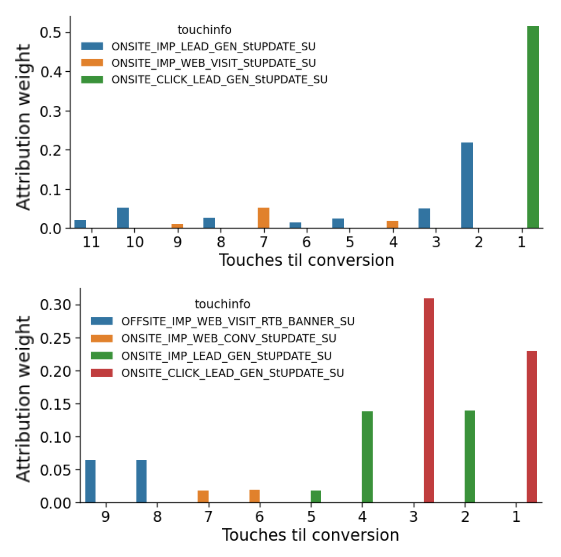}
    \caption{The attribution weights extracted from the model can provide granular insights. Here we show example attribution weights for two LinkedIn lead conversions, as a function of touches until conversion (x-axis) and touch type (hue). While in aggregate we see a time-decay effect (i.e., touches closer to the conversion tend to get more credit), the credit allocated to each touch in a given path can vary. For example, in the bottom path, the initial blue touches receive relatively more credit than the next few touches, and the red "click" three touches before the conversion receives the highest credit.
}
    \label{fig:lms_attribution_weights_ex}
\end{figure}

Touchpoint-level attribution from LiDDA provides very granular insights, where we see varying credit assignment to earlier touchpoints. In the provided examples in Figure~\ref{fig:lms_attribution_weights_ex}, touches closer to the conversion tend to have higher weights; however, the weight assigned to a specific touchpoint can vary, influenced by factors like the objective type, ad format, interaction type (impression, click), ad creative, days before conversion, etc. Using these types of signals, the model can learn to allocate credit across all touchpoints in the converting paths. This data can then be surfaced to advertisers on LinkedIn's Ad platform at different levels of aggregation, in a variety of ways, including in the Campaign Manager Reporting table (i.e., showing the number of conversions attributed to each campaign, creative, etc.), and in a plot in Measurement Insights (showing a time series of conversions attributed to different categories of campaigns).

\clearpage

\end{document}